\documentclass[letterpaper, 10 pt, conference]{ieeeconf}  

\IEEEoverridecommandlockouts                              
\usepackage{mathpazo} 
\usepackage{algorithm}
 \usepackage{algorithmic} 
 \usepackage{amsmath} 

\usepackage{graphicx}

\title{\LARGE \bf
Attention-Aware Generative Adversarial Networks (ATA-GANs)}

\author{ \parbox{6 in}{\centering Dimitris Kastaniotis, Ioanna Ntinou, Dimitrios Tsourounis, \\ George Economou and Spiros Fotopoulos, \\
         Department of Physics\\
          University of Patras\\
         Patras 26504, Greece \\
         {\tt\small dkastaniotis@upatras.gr}}
         }

\begin{document}

\maketitle
\thispagestyle{empty}
\pagestyle{empty}

\begin{abstract}


In this work, we present a novel approach
for training Generative Adversarial Networks (GANs). Using the attention maps produced by a Teacher-
Network we are able to improve the quality of the generated
images as well as perform weakly object localization on the generated images. To this end, we generate images of HEp-2 cells captured with Indirect Imunofluoresence (IIF) and study the ability of our network to perform a weakly localization of the cell. Firstly, we demonstrate that whilst GANs can learn the mapping between the input domain and the target distribution efficiently, the discriminator network is not able to detect the regions of interest. Secondly, we present a novel attention transfer mechanism which allows us to enforce the discriminator to put emphasis on the regions of interest via transfer learning. Thirdly, we show that this leads to more realistic images, as the discriminator learns to put emphasis on the area of interest. Fourthly, the proposed method allows one to generate both images as well as attention maps which can be useful for data annotation e.g in object detection.

\end{abstract}

\section{INTRODUCTION}
Over the recent years there is a great progress in the area of generative models. In particular, Generative Adversarial Networks (GANs) \cite{goodfellow2014}, proposed a radically new approach for training a  generative model. This formulation allows one to capture high dimensional and complex distributions efficiently using a minmax game between two Neural Networks - one that generates samples and one that evaluates them as real or fake. In this game and for image data, the generator learns the mapping from a low dimensional space to the space of images, allowing one to sample images from the high-dimensional distribution by sampling from a lower-dimensional distribution. These models have been applied in many computer vision tasks like  human brain decoding \cite{spampinato2017} or realistic images synthesis \cite{Shrivastava}. Furthermore, in tasks where data are limited (e.g. biomedical datasets) it can be used for data augmentation.

\par
One major limitation of current deep neural network models is the lack of a mechanism for knowledge transfer between tasks. Transfer learning aims to address this limitation by suggesting novel ways for transferring the experience between different tasks \cite{caruana1997} or even models\cite{hinton2014}. However, these works focus on transferring the knowledge of a network with respect to the ability of mapping images to a number of categories. Recently, authors in \cite{komodakis}, following a Teacher-Student approach, attempted to transfer the attention of a large network to one with fewer parameters by incorporating loss functions between the Teacher and the Student intermediate layers.  

\par
It has also been observed, that intermediate layers of CNNs act as weak object detectors \cite{Cinbis2015,Oquab2014,Oquab2015}. This occurs naturally as the filters, especially in higher layers, tend to synthesize objects. Authors in \cite{CAM} inspired by this observation, have shown that global average pooling layers can be used in order to create Class Activation Maps (CAMs).

\par
In this work, we focus on GAN  discriminator’s inability to locate the regions of interest when trying to learn how to discriminate real versus fake images. Specifically, we investigate if a discriminator uses patterns within areas of interest of an image. Our findings indicate that discriminators in a regular GAN fail to locate the object efficiently- thus evaluating regions that are not always of much interest. In this context, we propose a novel formulation for training GANs, using a Teacher network in order to help the discriminator learn where to pay attention to. Results indicate that  our method both improves the quality of the generated image as well as it provides a weakly localization of the objects on the generated images.

\section{PROPOSED METHOD}
\subsection{Overview}
Our method is mainly inspired by two recent advances in computer vision. Firstly, authors in \cite{komodakis}, showed that small CNNs achieve better performance, if they learn to mimic behavior of attention maps in intermediate layers of bigger networks. To this end, the attention maps of intermediate layers of a larger network are transferred to the small network by attaching several loss functions in intermediate feature maps. Additionally authors in \cite{CAM}, showed that CNNs which use average pooling at the last layer can naturally deliver a specific type of attention maps, called Class Activation Maps (CAMs). These attention maps can be used to perform weakly object detection \cite{Oquab2014}. As it is shown in Figure ~\ref{fig:Soft-CAM}, a modified version of Class Activation Maps (CAMs) can be used to help the discriminator pay more attention in regions of interest. The second one is the ability of GANs to generate realistic images by training a neural network to generate samples while a second one evaluates them as real or fake.

\begin{figure}
\centering
  \includegraphics[width=\linewidth]{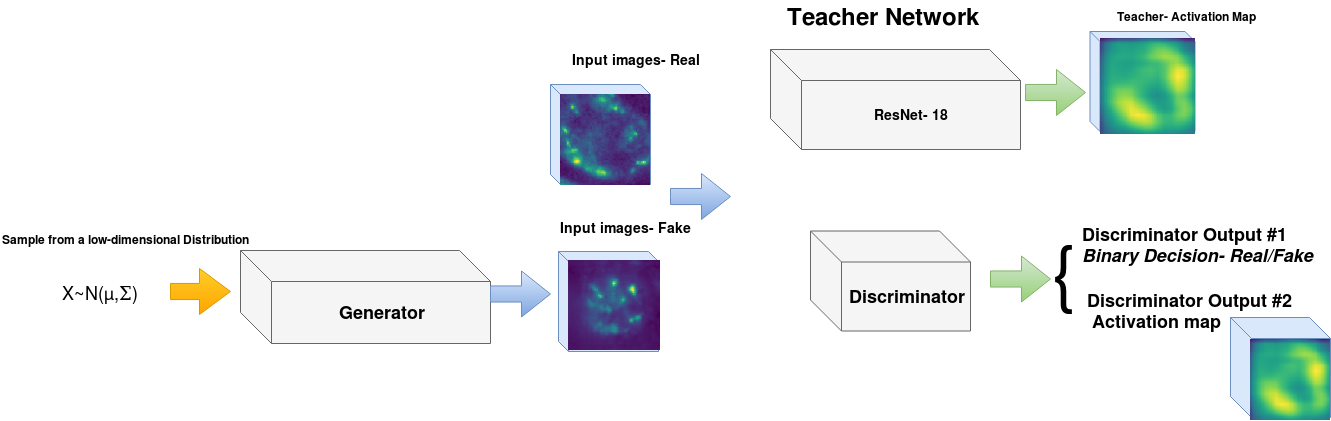}
  \caption{Overview of the proposed method}
  \label{fig:overview}
\end{figure}

\par
In this context, here we present a novel method that allows us to transfer the knowledge of a large network, originally trained to discriminate between different categories of HEp-2 cell images, into the discriminator of a GAN. To achieve this, we reformulate the method of CAMs \cite{CAM} in a way that allows us to incorporate the soft responses of all classes in the estimation of the CAM image. Our intuition is that there are some common features shared across HEp-2 cell image categories located in the same area, and hence, incorporating them according to a Soft-Max response allows us to proportionally consider patterns from all categories. The proposed method is named as Soft-CAMS (SCAMS). The Soft-CAMs , which are presented in the following section, will be used to guide the attention of a discriminator to better localize the cell. Therefore, we introduce an additional cost in the original GAN-minmax optimization problem. This cost is applied on the discriminator and enforces it to pay more attention into areas of interest (as they are defined by the Teacher network). The proposed method is summarized in Figure ~\ref{fig:overview}.

\subsection{Soft-Class Activation Maps}
The ability of Deep Convolutional Neural Networks to perform weak object detection in intermediate layers has been observed in a number of works \cite{Cinbis2015,Oquab2014,Oquab2015}.

Authors in \cite{CAM} applied global average pooling (which is used in many state-of-the-art deep neural network architectures) \cite{iandoloda2016,he2015} to perform weakly localization of objects in an image. Moreover, they presented a scheme, which allows one to determine the attention of a CNN on input image implicitly. In this work, we take advantage of their findings to extract the attention map from a large CNN and then use this to train the discriminator of the GAN. Furthermore, we present a modified version of the CAM which we call Soft-CAM (SC) that can be expressed as follows:

\begin{figure}
\centering
  \includegraphics[width=\columnwidth]{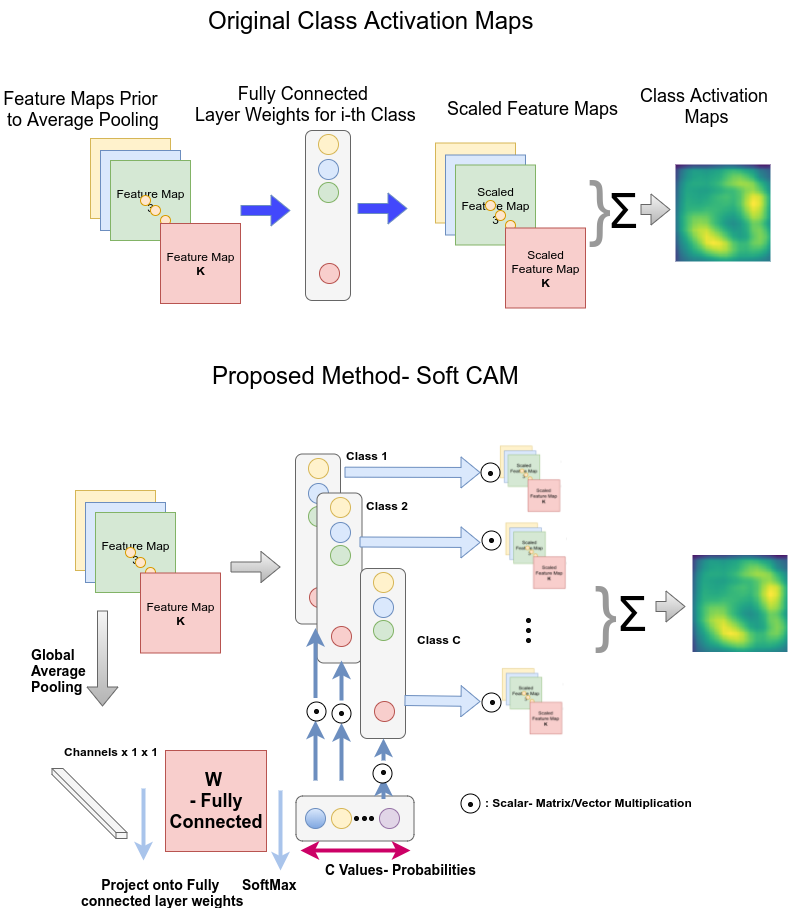}
  \caption{Soft-Max Activation Map}
  \label{fig:Soft-CAM}
\end{figure}

\begin{equation}
 SC =  \sum_{j=1,C}(SM(j)\cdot{ \sum_{i=1,K} FM_{i}\cdot W_{i,j}})
 \label{sc}
\end{equation}

In the previous equation $SM\in\mathbb{R^C}$ is a vector containing Soft-Max responses of the classifier, $K$ is a scalar equal to the number of channels before global average pooling and $C$ is a scalar equals the number of categories. This vector allows us to weight the contribution of each feature map in the Soft-CAM computation for each category proportionally to the classification probability. Index $j$ iterates over the number of categories and index $i$ iterates over the number of available feature maps. In total, there are $K$ feature maps and $C$ categories. Each feature map is a two dimensional image. Matrix $W \in \mathbb{R^KC}$ allows us to scale each \textit{i-th} feature map with a weight associated with the \textit{j-th} category. The Soft-CAM  is a (single channel) two-dimensional image with the same dimensions as the feature maps. This image sums proportionally the contribution of each class to the computation of CAM. It is important to notice that for the discriminator network, we have a single class while for the Teacher-Network we have more than one classes. One major advantage of the proposed scheme is that it allows one to naturally incorporate class information in the discriminator, without modifying the formulation.
\par

Figure ~\ref{fig:Soft-CAM}, displays a graphical illustration of the computation of Soft-CAM presented in Eqn. \ref{sc} as well as a comparison with the regular CAM computation.

\subsection{Attention transfer from an Expert Network}
As we have already claimed, the discriminator of a regular GAN, is unable to efficiently locate the region of interest (location of cell). This is also demonstrated in Figure ~\ref{fig:Soft-CAM of the discriminator}. In order to help the discriminator better localize the cell, here we are incorporating a large network (ResNet-18)\cite{he2015}, trained on a large corpus of HEp-2 cell images \cite{PR2014} which was able to perform a state-of-the-art weak localization of the cell inside image as shown in Figure ~\ref{fig:Soft-CAM of Teacher}.

\par
\begin{figure}
\centering
  \includegraphics[width=0.60\linewidth]{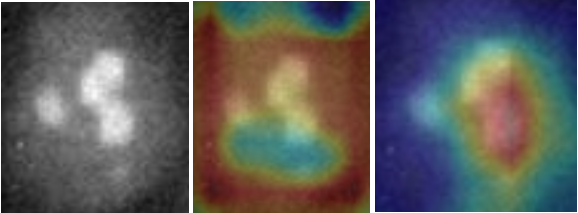}
  \caption{Left: The cell image. Middle: The attention map of a SqueezeNet drawn on top of a cell image. Red color indicates more contribution in the classification result. Right: The attention map of a ResNet.}
  \label{fig:SmallNet_TeacherNet}
\end{figure}

We also claim that this network should have enough capacity to learn the mapping between the images and the class categories. One way to ensure this, is to require that the network will achieve a human-level classification score. As we show in the experimental results, the proposed Teacher-Network is able to surpass human-level error. On the contrary, as we show in Figure ~\ref{fig:SmallNet_TeacherNet}, a SqueezeNet \cite{iandoloda2016} network (image in the middle) is not able to perform a robust weak localization. Also, in experiments we performed we noticed that SqueezeNet \cite{iandoloda2016} was not able to surpass human level error.

\subsection{Train GANS with Attention (ATA-GANS)}

In this section we first show that a discriminator in a GAN whilst being able to separate real from fake images, is not able to localize the region of interest of the cell. Then we present the original GAN minmax problem and our modified version that incorporates an attention cost which helps the discriminator to better localize the areas of interest. 
\par
The ability of the discriminator to locate the object of interest appropriately (here the cell) is very important. More specifically the discriminator might learn to classify images as fake by focusing on particular patterns in the boundaries of the image or even might focus on patterns that are not related to the information that is valuable for classification.  
\par
In the following Figure ~\ref{fig:Soft-CAM of the discriminator}, we show the Soft- Class Activation Maps (SCAMs) of two images- one real and one fake. It is obvious that the attention of the discriminator spreads across the whole image and does not focus on the area of interest- which in this case is the area where the cell is located.

\begin{figure}
\centering
  \includegraphics[scale=0.30]{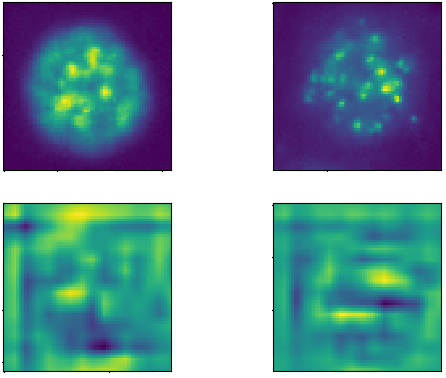}
  \caption{Soft-CAM of the discriminator of a regular GAN. Top-Left: A real image, Top-Right: A generated image. Bottom-Left: The attention map of the real image. Bottom-Right: The attention map of the generated image.}
  \label{fig:Soft-CAM of the discriminator}
\end{figure}

\begin{figure}
\centering
 \includegraphics[scale=0.2]{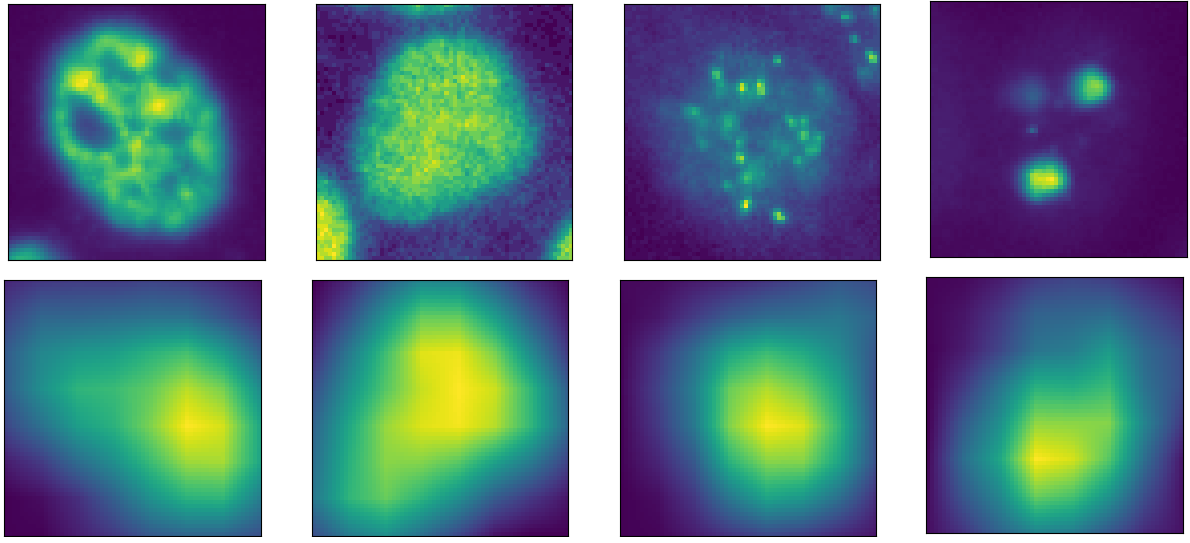}
\caption{Teacher Network: Soft-Class Activation Maps from Res-Net-18, trained on HEp-2 cell images. Top Row: Cell Images. Bottom Row: Soft-Class Activation Maps.}
 \label{fig:Soft-CAM of Teacher}
\end{figure}

As we have shown previously and depicted in Figure ~\ref{fig:Soft-CAM of the discriminator} the discriminator of a GAN,is unable to efficiently locate the region of interest (location of cell). In order to help the discriminator better localize the cell, we are incorporating a large network (ResNet-18) \cite{he2015}, trained on a large corpus of HEp-2 cell images \cite{PR2014} which is able to perform state-of-the-art weak localization of the cell inside the image as depicted in Figure ~\ref{fig:Soft-CAM of Teacher}. At the bottom we have the input images and at the top we have the Soft-Class Activation Maps.

\par
\begin{figure}
 \includegraphics[width=\columnwidth]{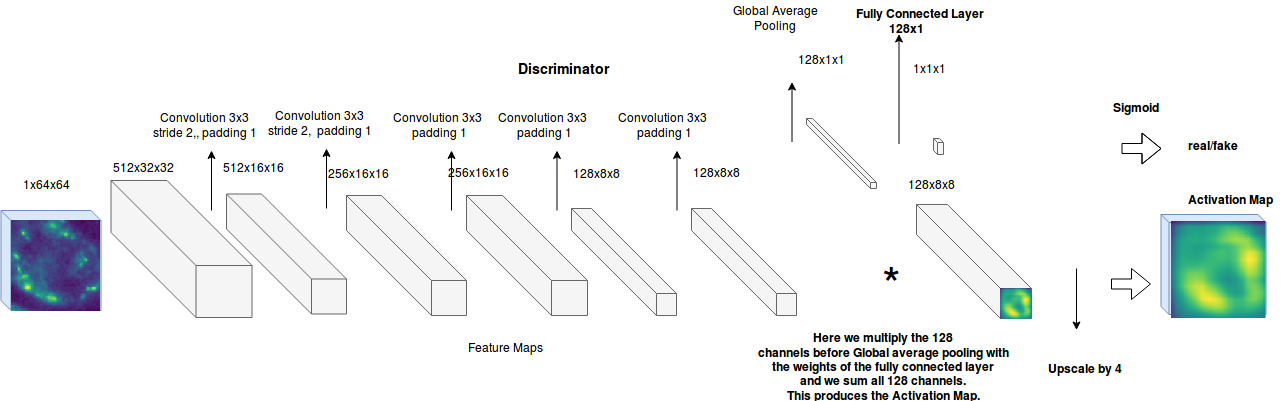}
 \caption{ The architecture of discriminator. The network produces a binary decision characterizing the input images as genuine or not, as well as a second output the S-CAM  }
\label{fig:Discriminator}
\end{figure}

During the training process, input images (both real and fake ones) are passed to the Teacher-Network as well as to the discriminator (see Figure ~\ref{fig:overview}). It is obvious, that the discriminator has to be implemented in a way that will produce these Soft-CAMs. In this context in Figure ~\ref{fig:Discriminator} , we present a version of the discriminator that produces two outputs. Firstly, it provides the binary decision of whether the input image is real or fake, and secondly, it produces the Soft-CAM image. 

The output of the teacher network together with the Soft-CAM output of the discriminator is then used in order to compute the following Mean-Square-Error (MSE) loss:
\begin{equation}
L_{SCAM} =    \frac{1\hfill}{2} (T_{SCAM}- D_{SCAM})^2
\label{eq:loss_attention}
\end{equation}

In the previous equation $T_{SCAM}$ is the attention map produced by the Teacher network presented at the top of Figure ~\ref{fig:overview} and $D_{SCAM}$ is the attention map produced by the discriminator and presented at the bottom of Figure ~\ref{fig:overview}.

However, this loss function is computed twice for the training of the discriminator, one for the real image and one for the generated image. Consequently, the same procedure is repeated for the generated image and the total loss is computed as 
\begin{equation}
L_{SCAM_{total}}={L_{SCAM_{real}}}+{L_{SCAM_{fake}}}
\label{eq:loss_attention_sum}
\end{equation}

This loss is being added to the Discriminator loss function only when training the discriminator. For the training of the Generator network we didn’t incorporate and attention loss as 
our main intuition is that we can improve the generator by allowing an expert network teach the discriminator to pay attention. The optimization problem is therefore formulated as follows:

 \begin{equation}
 \begin{split}
min_G max_D V(D,G)  =  \\ \mathbb{E}_{x \sim p_{data}} 
\label{eq:loss_attention_sum_both}
[\log(D(x))] +\\ \mathbb{E}_{z \sim p_{z}} [\log(1- D(G(x)))] 
\end{split}
\end{equation}

While we also add the minimization of the MSE criterion of Equation ~\ref{eq:loss_attention_sum} on the discriminator update step.


\section{Experimental Results}
 
Here we present some experimental results demonstrating the quality of generated images as well as the ability of the proposed scheme to provide both realistic new samples as well as weakly annotations. For this, in order to construct the Teacher network, we used ResNet-18 \cite{he2015}, which we trained a  on the task of HEp-2 cell image  classification using the dataset presented in \cite{PR2014}. This network achieved a remarkable performance on the dataset of HEp-2 cell images \cite{HEP2012} (76.28\% accuracy)  surpassing human level performance (73.3\%) and approaching the top performance (78.1\%) achieved in \cite{kastaniotis2017}. Also, this network was able to efficiently locate the object of interest on images as shown in Figure ~\ref{fig:Soft-CAM of Teacher}.
\par
\begin{figure}
  \includegraphics[width=\columnwidth]{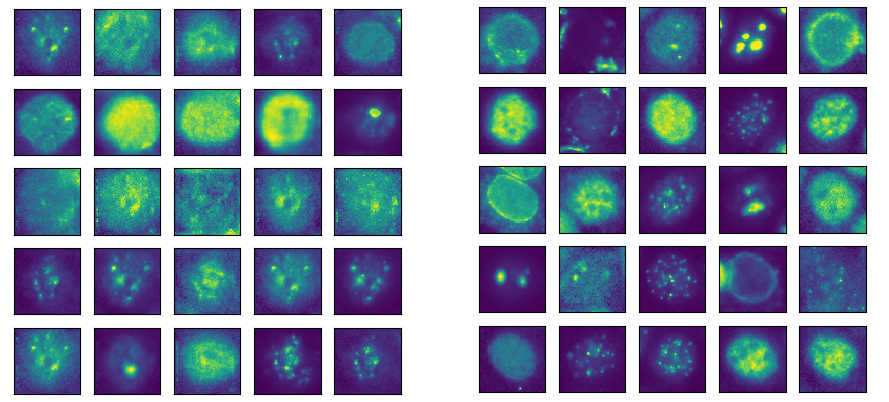}
  \caption{ Left: Generated Images, Right: Real Images }
  \label{fig:LeftGeneratedRightReal}
\end{figure}

Then we used this Teacher-network in order to train our discriminator by introducing one extra cost in the discriminator as shown in Equation ~\ref{eq:loss_attention_sum_both} during the minmax game between the generator and the discriminator. In Figure ~\ref{fig:LeftGeneratedRightReal} we provide some results of the generated images one the left side, and some real images on the right side. Our approach clearly generated very realistic images and most importantly, we verified that incorporating an  attention loss does not affect not affect negatively the quality of the generators images.

\par
\begin{figure}
  \includegraphics[width=\columnwidth]{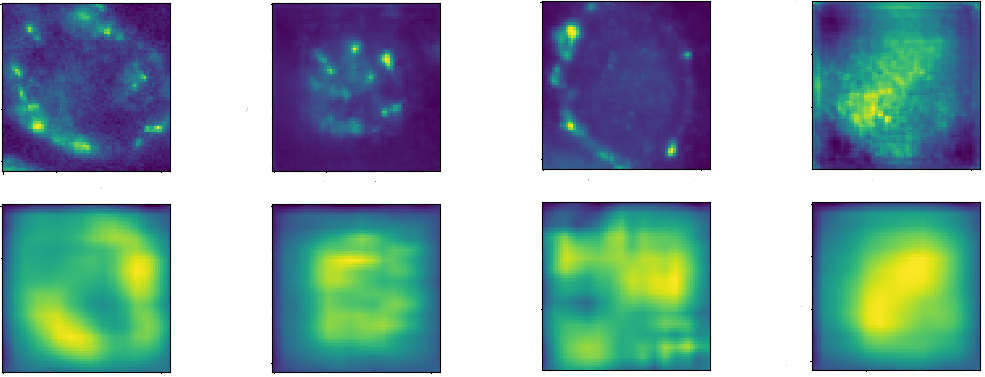}
  \caption{ Discriminator attention maps for input output pairs after training with our method. These images provide a weak localization of the cell area. On the left, we have two generated images (bottom row) and their Soft-CAMs (top row).On the right, we have two real images (bottom row) and their Soft-CAMs (top row).}
  \label{fig:AttentionMapsOurs}
\end{figure}

Regarding the ability of the proposed scheme to perform a weakly localization, in Figure ~\ref{fig:AttentionMapsOurs} we provide some Soft-CAMs from both generated as well as real images. As compared to Figure ~\ref{fig:Soft-CAM of the discriminator}, our modified minmax game was able to improve the attention of the discriminator significantly.

\par
Moreover, as we are interested in the ability of the network (generator) to generate realistic image and the discriminator to perform weak object localization, we provide some input-output pairs from both read as well as generated images. Results in the Figure ~\ref{fig:AttentionMapsOurs} verify that the proposed scheme can be used to both generate images as well as to provide weakly annotations.

\section{CONCLUSIONS}

In this work we proposed a method which allows the discriminator network of a GAN to perform weakly localization of the objects of interest. In order to achieve this we proposed a Teacher-Student learning scheme as well as a novel type of Soft-Class Activation maps. This scheme allows the discriminator to generate weak annotation of the generated images which can be used for automatic annotation of generated images. In the future we plan to apply this scheme on datasets designed for object detection in order to generate both images as well as weak annotations. We plan to make the code publicly available after the publication of this work.

\addtolength{\textheight}{-12cm}   


\end{document}